\documentclass[10pt,journal,compsoc]{IEEEtran}
\usepackage[nocompress]{cite}
\usepackage[pdftex]{graphicx}
\usepackage{amsmath}
\usepackage{algorithmic}
\usepackage{array}
\usepackage[caption=false,font=footnotesize,labelfont=sf,textfont=sf]{subfig}
\usepackage{url}
\usepackage[colorlinks=true]{hyperref}
\usepackage{ragged2e}

\begin{document}

\title{TIC: Text-Guided Image Colorization}
\author{
  Subhankar~Ghosh,
  Prasun~Roy,
  Saumik~Bhattacharya,
  Umapada~Pal,
  and~Michael~Blumenstein
  \IEEEcompsocitemizethanks{
     \IEEEcompsocthanksitem  S. Ghosh P. Roy, and M. Blumenstein are with Faculty of Engineering and IT,University of Technology Sydney, NSW, Australia
    \IEEEcompsocthanksitem  U. Pal is with the Computer Vision and Pattern Recognition Unit, Indian Statistical Institute, Kolkata, India.
    \IEEEcompsocthanksitem S. Bhattacharya is with Indian Institute of Technology, Kharagpur, India.
  }
}
\IEEEtitleabstractindextext{
\begin{abstract}
\justifying{Image colorization is a well-known problem in computer vision. However, due to the ill-posed nature of the task, image colorization is inherently challenging. Though several attempts have been made by researchers to make the colorization pipeline automatic, these processes often produce unrealistic results due to a lack of conditioning. In this work, we attempt to integrate textual descriptions as an auxiliary condition, along with the grayscale image that is to be colorized, to improve the fidelity of the colorization process. To the best of our knowledge, this is one of the first attempts to incorporate textual conditioning in the colorization pipeline. To do so, we have proposed a novel deep network that takes two inputs (the grayscale image and the respective encoded text description) and tries to predict the relevant color gamut. As the respective textual descriptions  contain color information of the objects present in the scene, the text encoding helps to improve the overall quality of the predicted colors. We have evaluated our proposed model using different metrics and found that it outperforms the state-of-the-art colorization algorithms both qualitatively and quantitatively.
}
\end{abstract}

\begin{IEEEkeywords}
Image Colorization, text-guided generation, GAN
\end{IEEEkeywords}
}

\maketitle

\section{Introduction}\label{sec1}

Old legacy movies and historical videos are in black and white format. When the video was captured, there was no suitable technology to preserve color information. Black and white or grayscale images can be restored by new real-life colorization, which gives life to old pictures and videos. The main aim of colorization is to add color to a black and white image or grayscale image such that the newly generated image is visually appealing and meaningful. In recent years, based on generative adversarial networks (GANs) \cite{goodfellow2014generative}, a variety of colorization techniques have been proposed, and the state-of-the art performance has been reported on current databases \cite{Caesar2018COCOStuffTA,WelinderEtal2010,Deng2009ImageNetAL}. These colorization techniques differ in many aspects, such as network architecture, different types of loss functions, learning strategies, etc. However, the existing colorization processes mostly follow unconditional generation where the colors are predicted only from the grayscale input image. This might lead to ambiguous results as the prediction of color from a grayscale information is inherently ill-posed. To increase the fidelity in the colorization pipeline, we propose a text-guided colorization pipeline where some color descriptions about the objects present in the grayscale image can be provided as auxiliary conditions to achieve more robust colorized results.\\
The major contributions of our work are as follows.
\begin{itemize}
    \item We propose a novel GAN pipeline that exploits textual descriptions as an auxiliary condition. 
    \item We extensively evaluate our framework using qualitative and quantitative measures. In comparison with the state-of-the-art (SOTA) algorithms, we found that the proposed method generates results with better perceptual quality. 
    \item To the best of our knowledge, this is the first attempt to integrate textual information in the colorization pipeline to improve the quality of generation. The textual color description acts as an additional conditioning to increase the fidelity in the final colorized output.  
\end{itemize}
\begin{figure}[]
  \centering
  \includegraphics[width=0.9\linewidth ]{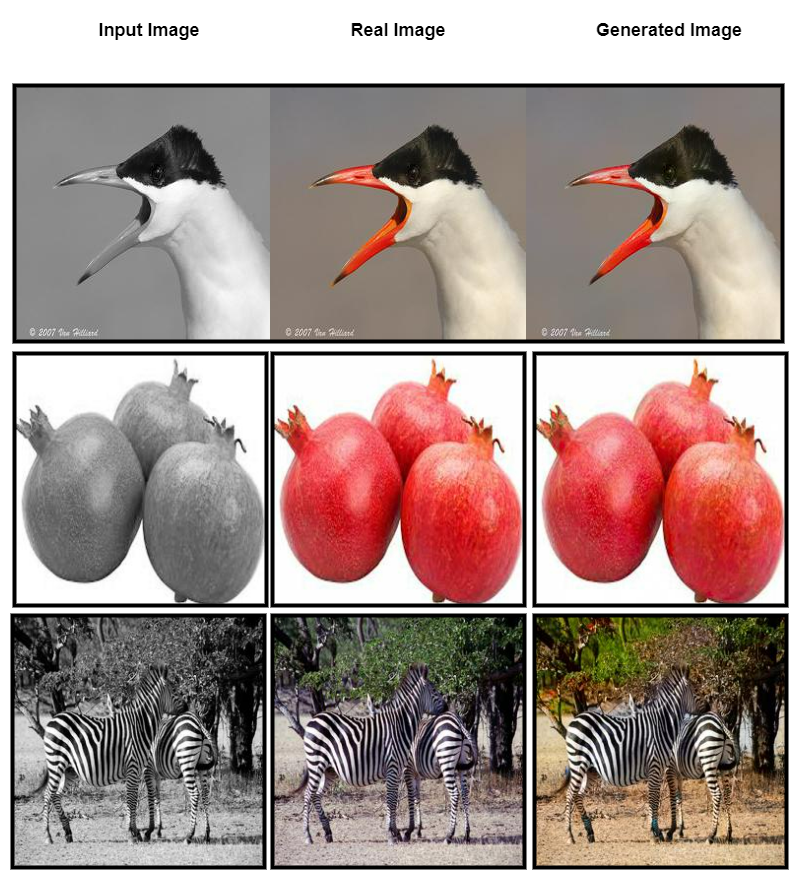}
  \caption{Images generated by the proposed algorithm: the first column indicates the input grayscale images; the second column shows the ground truth color images and the third column illustrates the respective colorized outputs of the proposed model.  }
    \label{fig:result}
\end{figure}

\begin{figure*}
  \centering
  \includegraphics[width=\linewidth]{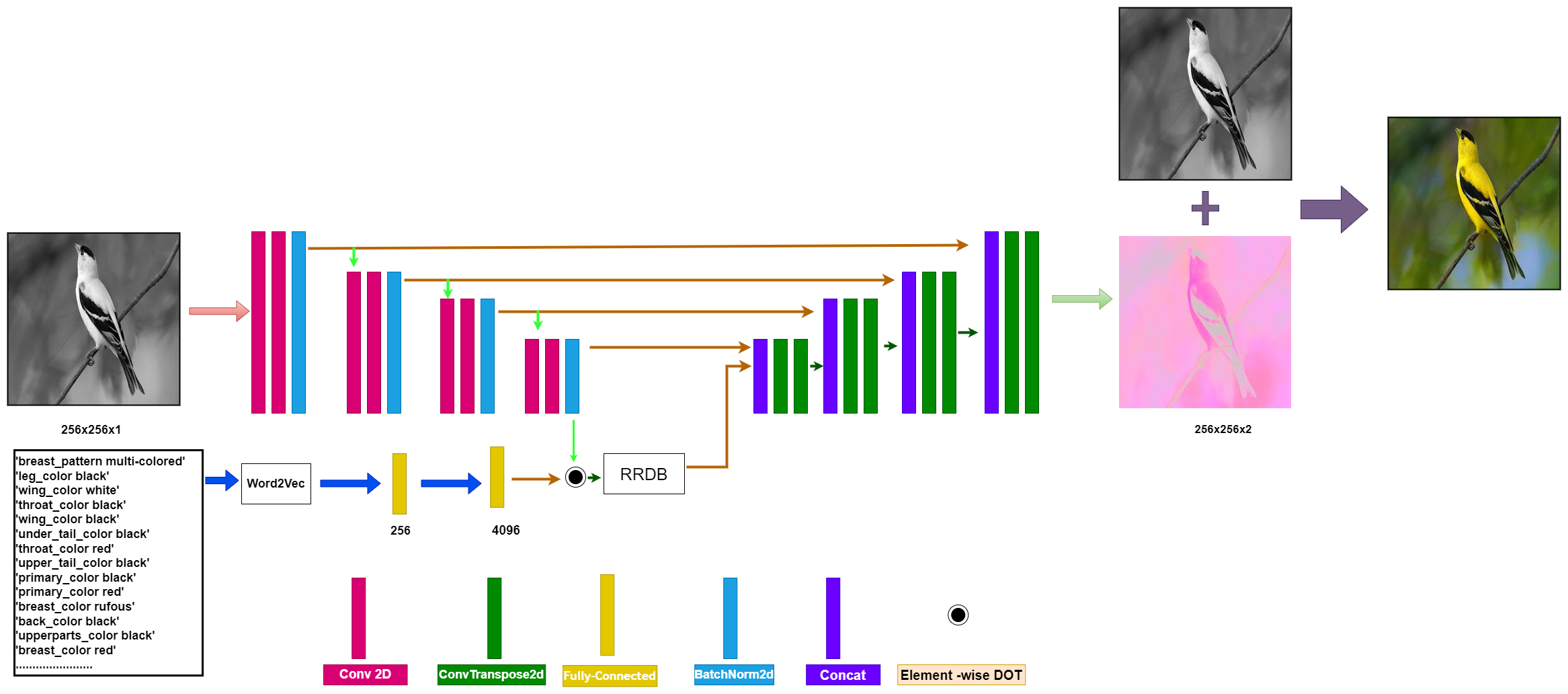}
  \caption{The block diagram of the proposed architecture. The network predicts the color components of the image which is combined with the intensity image to produce the final colorized image.}
    \label{fig:architecture}
\end{figure*}

\begin{figure*}
  \centering
  \includegraphics[width=\linewidth]{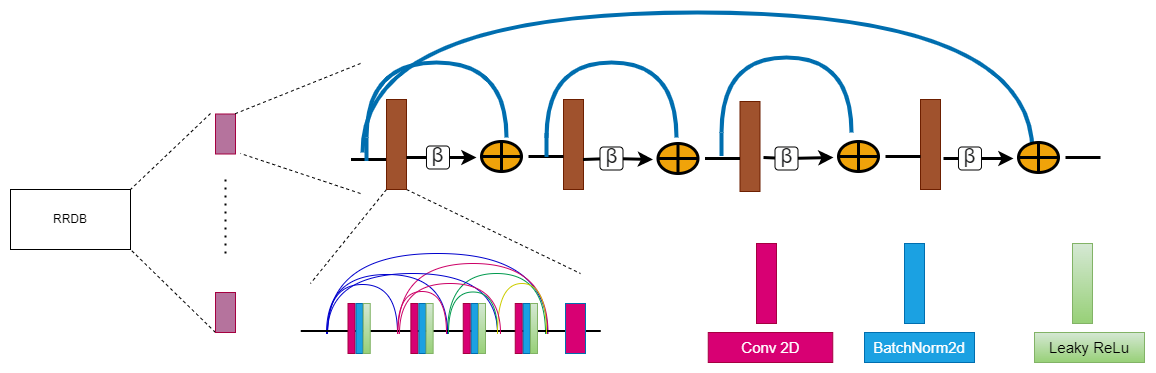}
  \caption{The block diagram of the Residual in Residual Dense Block(RRBD) architecture. }
    \label{fig:RRBD}
\end{figure*}

The rest of the paper is organized as follows. Sec. \ref{sec:relworks} introduces the SOTA colorization techniques. In Sec. \ref{sec:method}, we discuss our proposed colorization framework in detail. Sec. \ref{sec:exptdet} presents the experimental settings that are used to train and evaluate the pipeline. We present our results, and compare our proposed framework with the SOTA algorithms using qualitative and quantitative metrics in Sec. \ref{sec:results}. Finally, in Sec. \ref{sec:concl}, we conclude the paper by pointing out the overall findings of the proposed work, its limitations and future prospects.

\section{Related work}
\label{sec:relworks}
Image colorization methods have been the primary focus of significant research over the last two decades.  Most of these methods were influenced by conventional machine learning approaches \cite{Levin2004ColorizationUO,Huang2005AnAE,Bastos2013RUNTIMEGS}.  In the last few years, the trend has shifted to deep learning (DL)-based approaches due to the success of DL-based approaches in different fields \cite{anwar2020ColorSurvey,Wang2018GeneratingHQ,Tola2008AFL,Perazzi2016ABD}.  Recently, DL-enabled automatic image colorization systems  have  shown  an  impressive  performance  in the colorization task \cite{Zhang2016ColorfulIC,Carlucci2018DE2CODD,Levin2004ColorizationUO,Cheng2015DeepC,Wang2018GeneratingHQ,Bahng2018ColoringWW}.

Deep colorization \cite{Cheng2015DeepC} was the first network to incorporate DL for image colorization.  In training, five Fully Connected layers are used followed by ReLU activation with  the least-squares error as a loss function.  In this model, the first layer neurons depend on the features extracted from the gray scale patch.  The output layers have only two neurons, i.e., the U and V channel. During testing, grayscale image features are extracted from three levels, i.e., low-level, mid-level, and high-level. The sequential gray values, DAISY feature \cite{Tola2008AFL}  and semantic labeling are extracted at the low level, mid-level and high level, respectively to complete the task.

Deep depth colorization \cite{Carlucci2018DE2CODD} used a pre-trained ImageNet \cite{Deng2009ImageNetAL} network for colorizing  the  image  using  their  depth  information  of  RGB.  First,  the  network is  designed  for  object  recognition  by learning  a  mapping  from  the  depths to  the  RGB  channel. Pre-trained  weights  are  kept  frozen  in  the  network,  and this  pre-trained  network  is  merely  used  as  a  feature  extractor.\\
Wang \textit{et al.} \cite{Wang2018GeneratingHQ} proposed SAR-GAN to colorize Synthetic Aperture Radar images. The cascaded generative adversarial network is used as the underlying architecture. SAR-GAN was developed with two subnets; one is the speckling subnet, and another is the colorization subnet. The speckling subnet generates the noise-free SAR image, and the colorization subnet further processes it to colorize the images.
The speckling sub-network consists of 8 convolution layers with BatchNorms and element-wise division in the residual network. The colorization subnet utilizes an encoder-decoder architecture with 8 convolution layers and skips connection. The Adam \cite{Kingma2015AdamAM} optimizer is used for training the entire network. The SAR-GAN utilizes hybrid loss with l1 loss and adversarial loss.\\
The text2color \cite{Bahng2018ColoringWW} model consists of two conditional adversarial networks: the Text to Palette Generation network and the Palette-based colorization network. The Text to Palette Generation network is trained using the palette and text dataset. The text to Palette Generation networks generator learns the color palette from the text and identifies the fake and real color palette. Huber loss is used as a loss function in this network. The palette-based colorization network is designed using the U-NET architecture where the color palette is used as a conditional input to the generator architecture. The authors had designed the discriminator using a series of convo2d and LeakyRelu \cite{Maas2013RectifierNI} modules, followed by a fully connected layer to classify the colorized image as real or fake.

\section{Methodology}
\label{sec:method}
Image colorization aims to generate a  color image from a  grayscale image. Typically, deep learning tools use RGB images as ground truth for image generation. In the proposed method, RGB images are converted into the CIE LAB color space, where we need to find only the `A' and `B' channels instead of three channels of RGB. We converted the input text, containing the color information of the image, to a word vector using the word2vec \cite{Mikolov2013EfficientEO}. The size of the word vector is 256, and the input size of the image is $256\times 256$, which is the `L' channel of the LAB color space. We add the `L' channel with the AB channel of the image, which the Generator predicts, to reconstruct a fully colorized image. The discriminator signifies the visual authenticity of the image in a patch-based manner.

\subsection{Generator}
The idea of the proposed Generator is that the text color information is fused with the grayscale image ($L^i$) at the last downsample step of the network. The input L image is first resized to a fixed size of $256\times 256$. The overall generator has two pathways- an image pathway, through which the image information flows in the network, and the text pathway, through which the text color information flows as a conditional input. Both the pathways finally meets in the Residual in Residual Dense Block (RRDB). For the image path, each resolution level has two convolution layers. The down-sampling follows the last convolution by 2 to move to a new resolution. We use a $3\times 3$ kernel size with 64 filters in each convolution block. After each convolution, we perform batch-normalization \cite{Ioffe2015BatchNA}, and each convolution block has ReLU \cite{nair2010rectified} activation. We also process the text vector($S^i$) by two fully connected layers of sizes 256 and 4096. We resize the text features computed by the last fully connected layer to 1x64x64 and perform an element-wise dot product between the image features and the text features to impose a text-guided conditioning. The text conditioned features are then fed to a Residual in Residual Dense Block (RRBD) \cite{wang2018esrgan} before forwarding to the expanding part of the generator. The RRBD block consists of several dense layers with skip connections. The output of each  dense block is scaled by  $\beta$ before feeding it to the next dense block. Each Dense block consists of a convolutional layer, followed by BN and leaky ReLU activation with the residual connection. As shown in Fig. \ref{fig:RRBD}, skip connections are introduced to tackle the problem of a vanishing gradient. The output of the RRBD block  is used as input to the convTranspose2d layer with a 64 filter. In the expanding pathway, we have three up-sampling oparations  that work in four different resolutions. To increase the feature information in the expanding path, after each up-sampling layer, we concatenate the features available with the same resolution in the contracting path. The convolution blocks in the expanding path are similar to the convolution blocks at the contracting paths, and we decrease the number of filters by two as we move to the higher resolution. At the highest resolution, we apply two filters with kernel size 1x1 to generate the estimated $AB$ channel of the color image. At the end of the proposed network, we compute the color image by adding  the generated $AB$ and the input grayscale image($L^i$).  We illustrate the proposed Generator in Fig \ref{fig:architecture}.

\subsection{Discriminator}
For the colorization task, it is required that the discriminator can detect the local quality of a generated colorized image. Thus, we use the PatchGAN \cite{Isola2017ImagetoImageTW} Discriminator $D$ to judge the quality of the generated image. The discriminator penalizes the generated structure at the patch level resulting in a high-quality single level generation. We stack the grayscale image ($L^i$) with either a target image ($T^i$) or with a estimated image ($E^i$) where  $T^i$ and $E^i$ are the $AB$ channel of the color image. The ($L^i$,$T^i$) stack is labeled as real and the ($L^i$,$E^i$) stack is labeled as fake. In this way, we enforce discrimination on image transition rather than the image itself. In our model, the Patch discriminator takes a three-channel input dimension $256\times 256$. The discriminator has three convolution blocks with 64, 128 and 256 filters, respectively, in each block with filter dimension $4\times 4$. In the first two convolution blocks, the filter has stride 2, whereas, for the last two blocks, we used stride 1x1. Each convolution layer is followed by batch-normalization \cite{Ioffe2015BatchNA} and leaky-ReLU \cite{Maas2013RectifierNI} activation. After the convolution blocks, we apply one filter of kernel size 4x4 with stride 1 to compute the final response. The average of the final response is the output of the discriminator.
\subsection{Training}
As mentioned in \cite{Isola2017ImagetoImageTW}, the PatchGAN discriminator focuses more on the high frequency information. Thus to keep the fidelity of low frequency information in the colorized image, we used $L_1$ loss in the generator G which is calculated as
\begin{equation}
\mathcal{L}^G_1=\|E^i-T^i\|_1=\|G(L^i,S^i)-T^i\|_1
\end{equation}
As we have trained the generator in an adversarial manner, we define the adversarial or the GAN loss of the generator and the discriminator as:

\begin{equation}
\mathcal{L}^G_{GAN}=\mathcal{L}_{BCE}(D(L^i,G(L^i,S^i)),1) 
\end{equation}
\begin{equation}
\mathcal{L}^D_{GAN}=\mathcal{L}_{BCE}(D(L^i,T^i),1)\nonumber+\mathcal{L}_{BCE}(D(L^i,G(L^i,S^i)),0)
\end{equation}

where ${L}^G_{GAN}$ and ${L}^D_{GAN}$ denote adversarial Generator loss adversarial discriminator loss, respectively. To increase the visual quality of the image, we use perceptual loss to train the generator. 

\begin{equation}
\mathcal{L}_{p_\rho}^G=\frac{1}{h_\rho w_\rho c_\rho}\sum_{x=1}^{h_\rho}\sum_{y=1}^{w_\rho}\sum_{z=1}^{c_\rho}\|\phi_\rho(E^i)-\phi_\rho(T^i)\|_1
\end{equation}

where $\mathcal{L}_{p_\rho}^G$ is the perceptual loss computed at the $\rho^{\text{th}}$ layer, $\phi_\rho$ is the output from  the $\rho^{\text{th}}$ layer of a pretrained VGG19 \cite{Simonyan2015VeryDC} model, and $h_\rho$, $w_\rho$ and $c_\rho$ are the height, width and the number of channels at that layer, respectively.\\
The total generator loss $\mathcal{L}^G$ can be defined as
\begin{equation}
\mathcal{L}^G= arg \min_{G} \max_{D}\text{ }\lambda_1\mathcal{L}^G_{GAN}+\lambda_2\mathcal{L}^G_{P_4}+\lambda_3\mathcal{L}^G_1
\end{equation}

\section{Experimental Details}
\label{sec:exptdet}
We use the PyTorch framework to build the model, and perform our experiments. We use the Adam \cite{kingma2014adam} optimizer to train both the generator and discriminator up to 350K iterations with  $\beta_1=0.5$ and $\beta_2=0.999$. The learning rate is $1\times 10^{-4}$ with a decay of 0. All the leaky-ReLU activations have negative slope coefficients of 0.2. We select $\lambda_1=1$, $\lambda_2=1$ and $\lambda_3=1$.\\

While training the discriminator $D$, we concatenate $L^i$ with either $T^i$ or $E^i$ and give that as a input. Both $D$ and $G$ are trained iteratively, $ i.e., $ we keep $D$ fixed while training G and vice versa. As the training process of GAN is highly stochastic, we store the network weights at the end of each iteration. At the time of inference, we drop the discriminator network and generate the A,B channels only using the generator network.
\subsection{Datasets}
\begin{figure}
  \centering
  \includegraphics[width=0.5\linewidth]{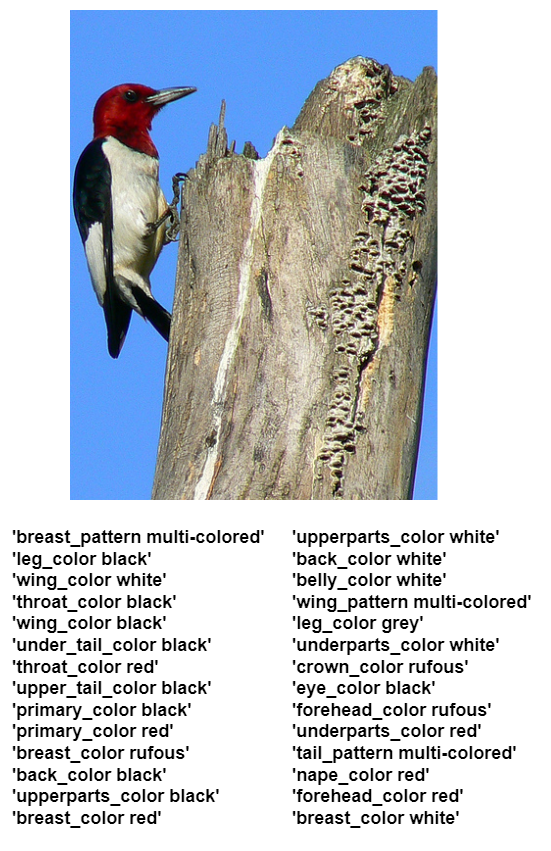}
  \caption{A typical example in our dataset: each sample contains a color image and corresponding color descriptions of the bird. To use the image while training the network, we first convert the color image to LAB color space, and use the `L' image as the input.}
    \label{fig:example}
\end{figure} 
To evaluate the performance of our model, we use three popular datasets, Caltech-UCSD Birds 200 \cite{WelinderEtal2010}, MS COCO \cite{Lin2014MicrosoftCC} and Natural color Dataset(NCD) \cite{anwar2020ColorSurvey}.\\
The Birds dataset contains 6032 bird images with their color information. We split the dataset into two parts (train, test). The total number of images for training is 5032, and the remaining images are used in the test set.

The total number of images in the NCD set are 730 fruit images. We use 600 images for training and the remaining 130 images for testing. We converted the class label to  one single color, like the tomato's class is converted into red and used as color information for training and testing. 

From the MS COCO \cite{Lin2014MicrosoftCC} dataset, we use 39k images for training and 6225 images for  testing. In COCO stuff \cite{Caesar2018COCOStuffTA}, the text description of the images  are available. In each text description, we find the sentence(s) related to color information of an object to provide it as auxiliary information. We collect all such sentences and use it as the final auxiliary information for the respective image.   

\section{Experimental Results}
\label{sec:results}
\begin{figure}
  \centering
  \includegraphics[width=\linewidth]{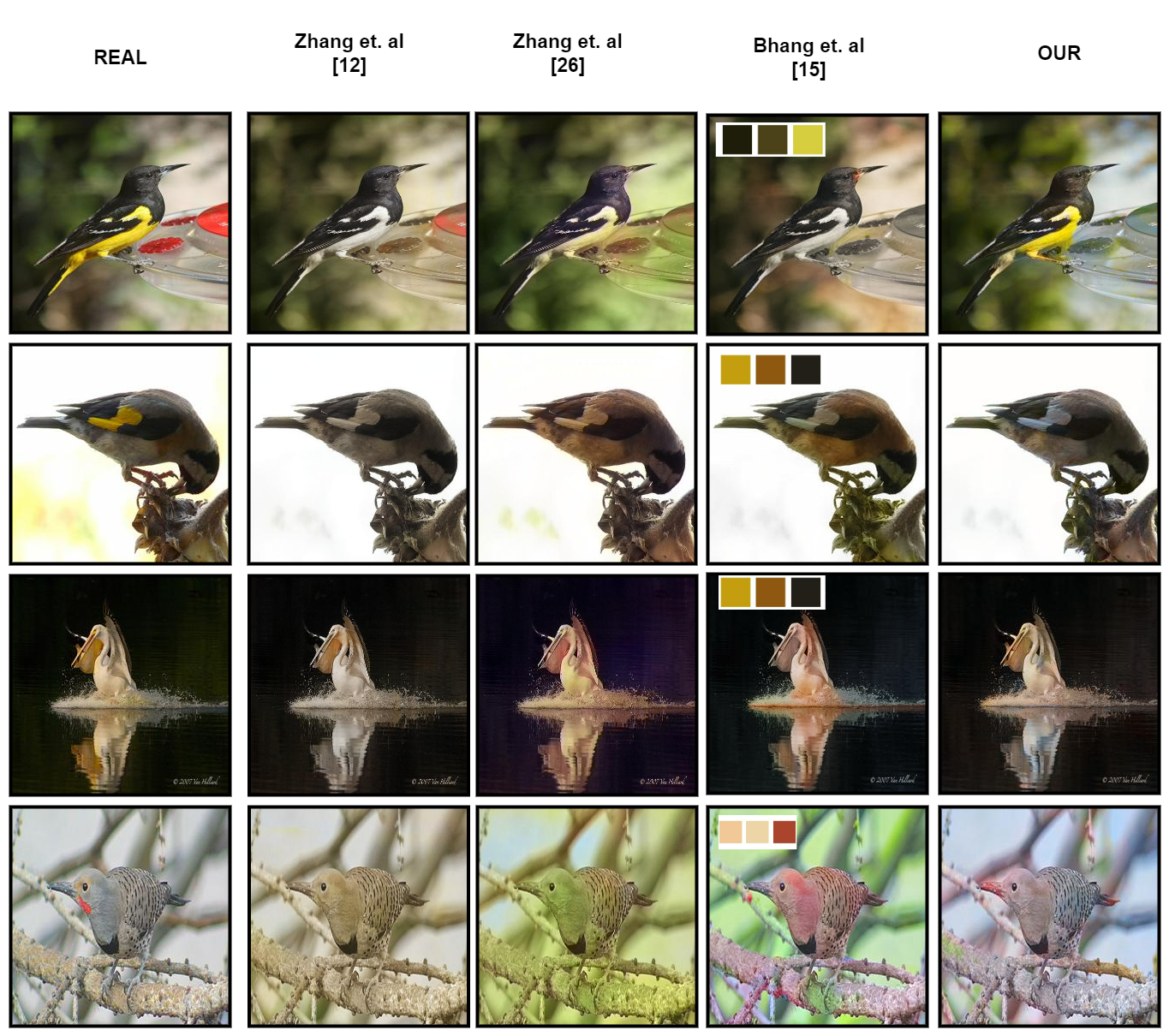}
  \caption{Qualitative comparison results: The first column contains ground truth images, the second column, third and fourth columns contain the results generated by the SOTA algorithms, and the last column shows the results generated by the proposed algorithm.}
    \label{fig:compare}
\end{figure} 
\begin{figure}
  \centering
  \includegraphics[width=.9\linewidth ]{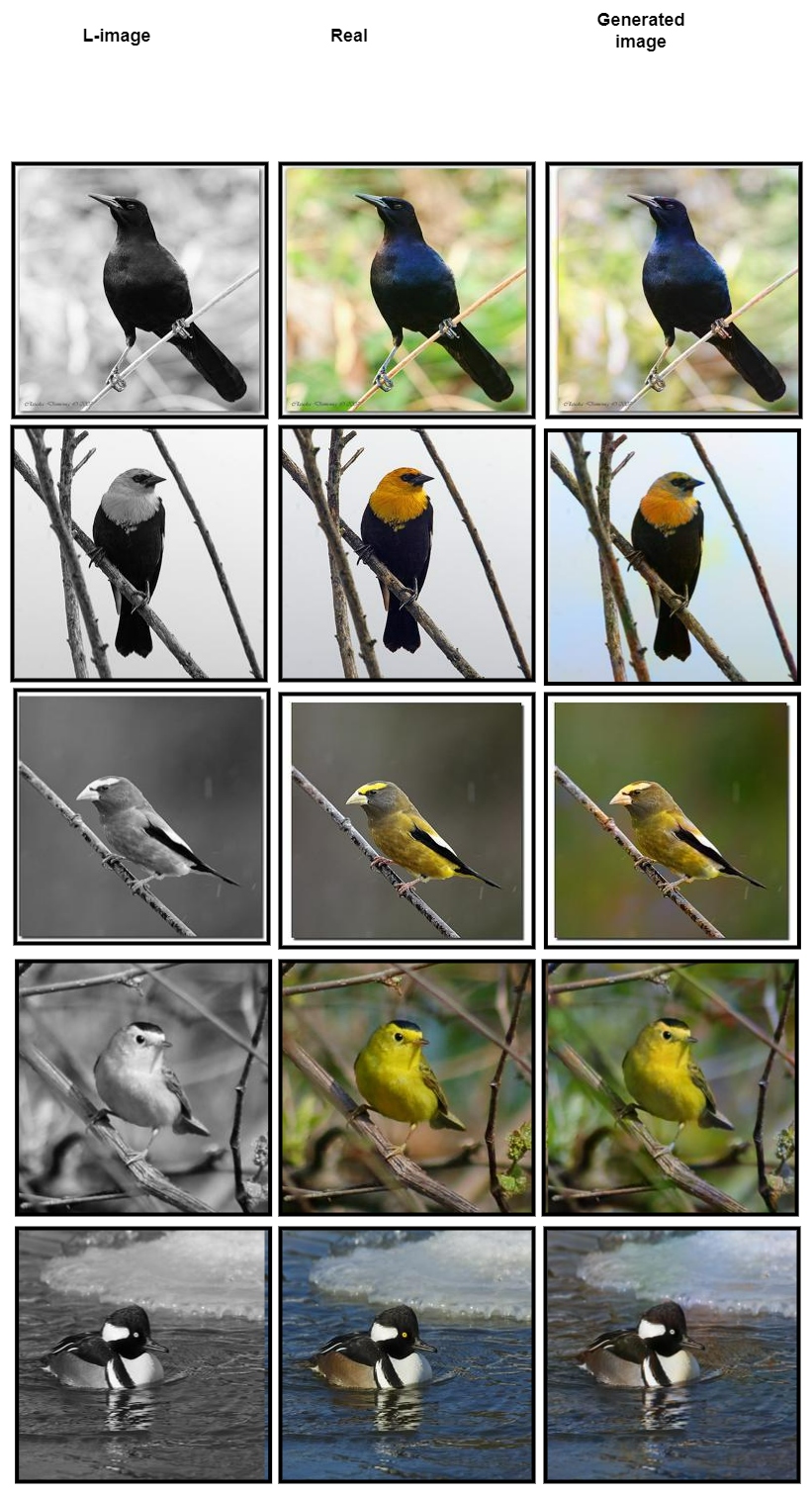}
  \caption{Images generated by the proposed algorithm from the Caltech-UCSD Birds 200 \cite{WelinderEtal2010}: the first column contains the grayscale images, the second column contains the ground truth images and the third column shows the colorized outputs of the proposed model.   }
    \label{fig:result1}
\end{figure}
\begin{figure}
  \centering
  \includegraphics[width=.9\linewidth ]{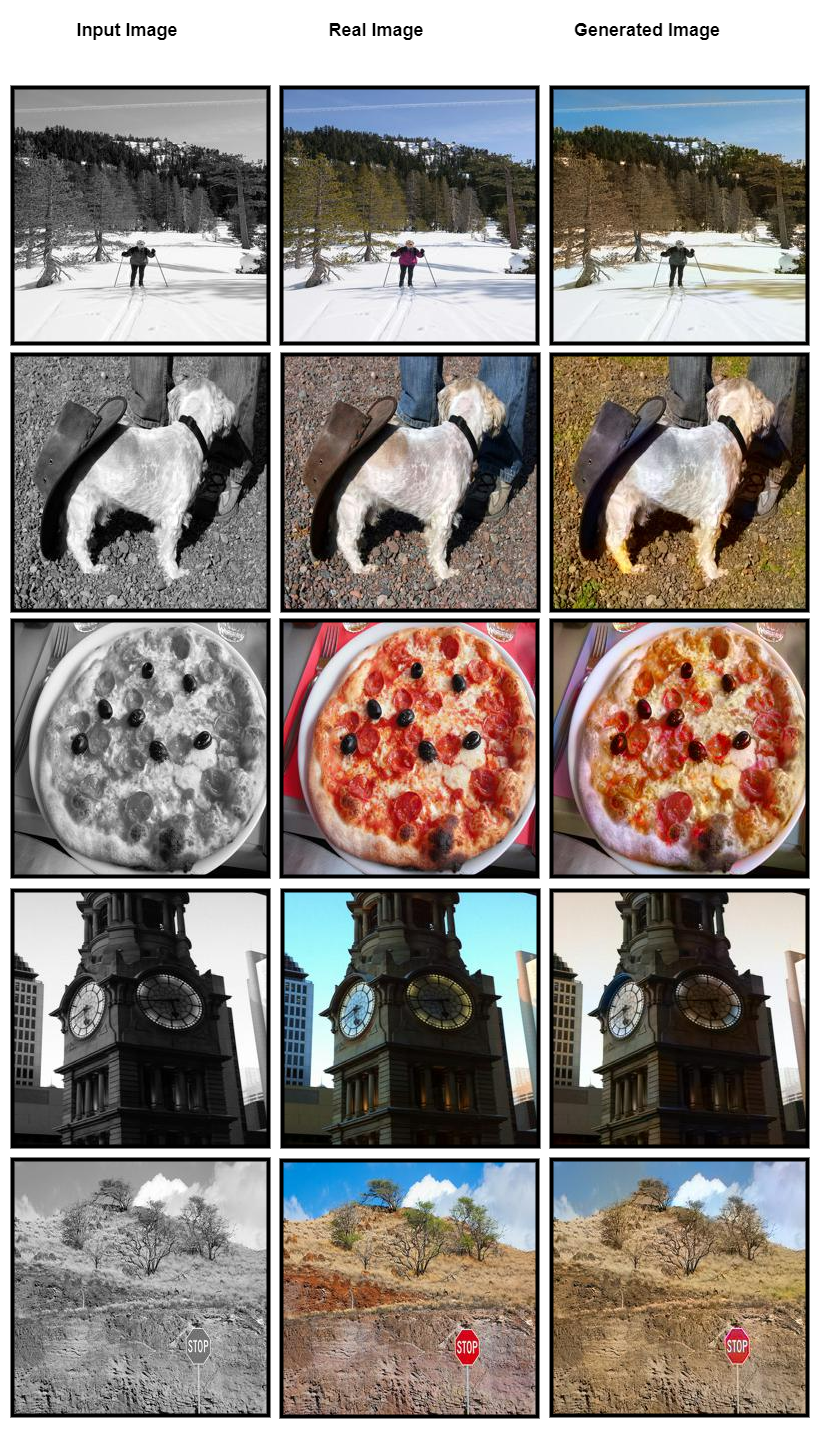}
  \caption{Images generated by the proposed algorithm form the MS COCO stuff \cite{Caesar2018COCOStuffTA} Dataset: the first column contains the grayscale images, the second column contains the ground truth images and the third column shows the colorized outputs of the proposed model.   }
    \label{fig:result2}
\end{figure}
\begin{figure}
  \centering
  \includegraphics[width=.9\linewidth ]{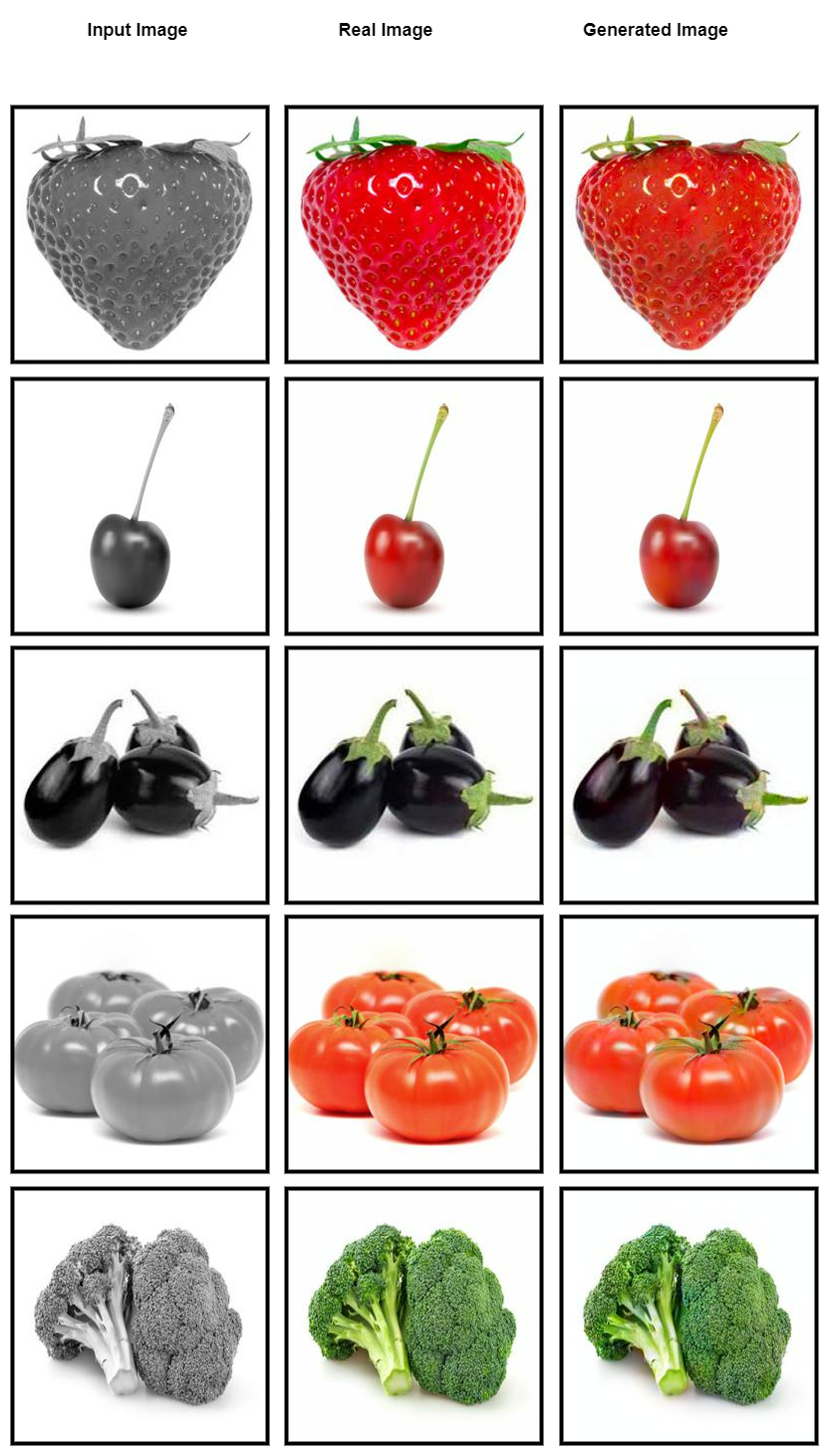}
  \caption{Images generated by the proposed algorithm from the NCD \cite{anwar2020ColorSurvey} Dataset: the first column contains the grayscale images, the second column contains the ground truth images and the third column shows the colorized outputs of the proposed model.   }
    \label{fig:result3}
\end{figure}

\begin{figure}
  \centering
  \includegraphics[width=\linewidth]{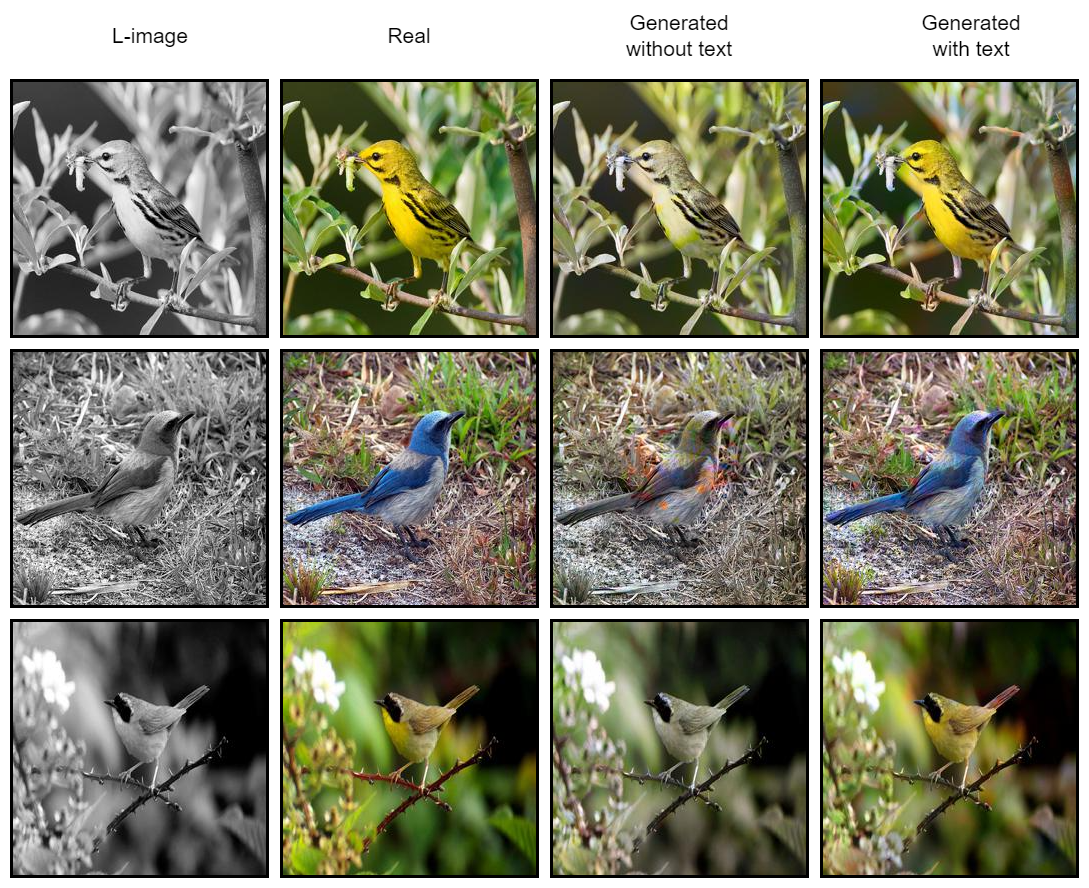}
  \caption{Validation of the importance of the textual encoding: first column contains the grayscale images, second column contains the ground truth images, the third and fourth columns show the results generated without and with the textual encoding, respectively. }
    \label{fig:ablation}
\end{figure}

\begin{figure}
  \centering
  \includegraphics[width=\linewidth]{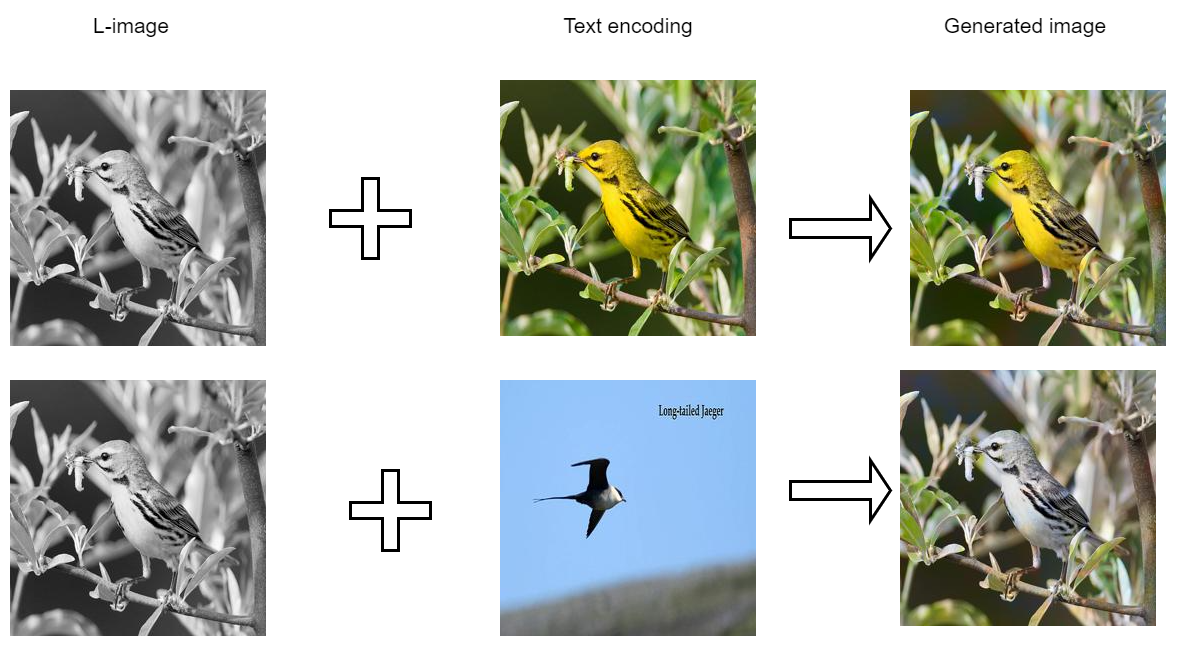}
  \caption{Recolorization: The first column shows the grayscale images, the second column shows the images whose textual descriptions are used as conditioning. The third column shows the final colorized images.  }
    \label{fig:text_ablation}
\end{figure}

\begin{figure}
  \centering
  \includegraphics[width=\linewidth]{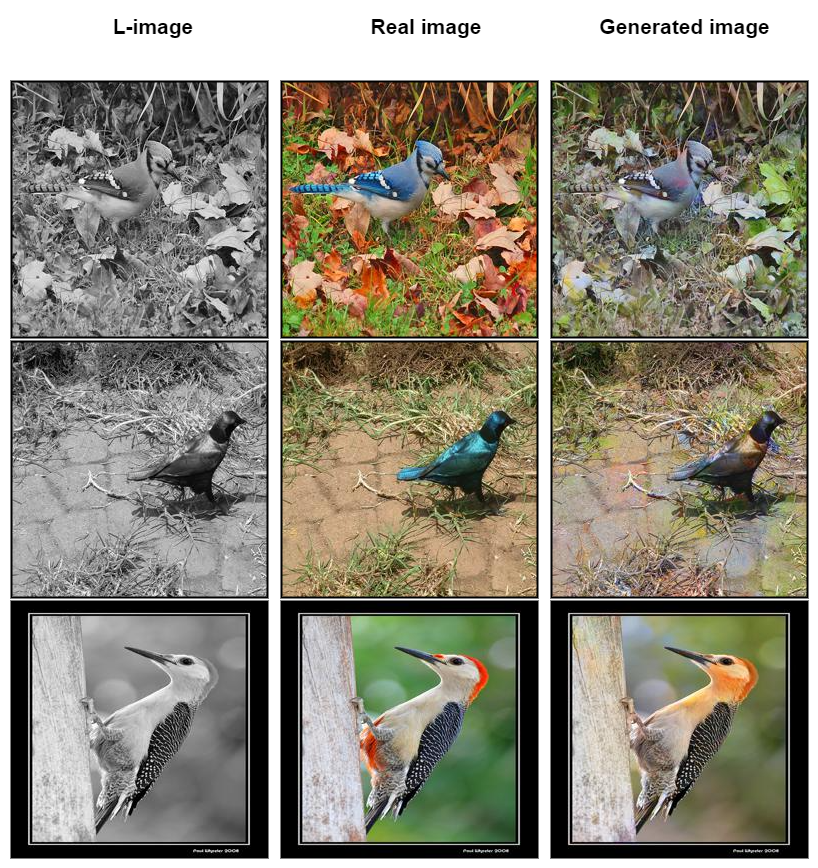}
  \caption{Some of the failure cases. }
    \label{fig:limitation}
\end{figure}

To understand the overall performance of the proposed framework, we performed an extensive set of experiments to evaluate the quality of the final colorized images. In Fig. \ref{fig:compare}, we compare our algorithm with  \cite{Zhang2016ColorfulIC}, \cite{Zhang2017RealtimeUI} and \cite{Bahng2018ColoringWW}. As shown in the figure, the proposed algorithm colorized the grayscale images with higher fidelity. Although the existing methods have colorized the grayscale images successfully, however the colors are often significantly different to the actual ground truth. The colorized images produced by the SOTA algorithms are also less colorful. As the proposed method utilizes the textual description as auxiliary information, our algorithm generates more realistic and colorful images from the respective grayscale input images. We have generated the color images from three different public databases. Fig. \ref{fig:result1} shows image samples from the UCSD Bird dataset \cite{WelinderEtal2010}. Fig. \ref{fig:result2} and Fig. \ref{fig:result3}  show samples from the MS COCO \cite{Caesar2018COCOStuffTA} dataset and the Natural Color Dataset \cite{anwar2020ColorSurvey}, respectively.  To validate the importance of the textual description further, we train a new model without using the textual information. As shown in Fig. \ref{fig:ablation}, without the textual conditioning, the proposed pipeline fails to colorize the grayscale images properly. In Fig. \ref{fig:text_ablation}, we also demonstrate that the textual description can be used for the recolorization task. In Fig. \ref{fig:text_ablation} (a), we have colorized the  grayscale image with the actual textual description of the ground truth. In Fig. \ref{fig:text_ablation} (b), we have kept the grayscale image unchanged and have used the textual description of a different image. It is observed that the proposed framework is able to follow the textual conditioning, and can produce significantly different colorized outputs from the the same grayscale image based on the textual encodings. \\

 \begin{table}[]
 \caption{Quantitative comparison among different colorization methods -- Zhang et. al \cite{Zhang2016ColorfulIC}, Zhang et. al \cite{Zhang2017RealtimeUI}, Bhang et. al \cite{Bahng2018ColoringWW} and our method.}
\centering
\begin{tabular}{lllll}
Method  & SSIM $\uparrow$  & PSNR $\uparrow$  & LPIPS (vgg) $\downarrow $  & LPIPS (sqz)$\downarrow$  \\\hline
Zhang et. al \cite{Zhang2016ColorfulIC} & 0.903          & 22.94      & 0.253       & 0.143     \\
Zhang et. al \cite{Zhang2017RealtimeUI} & 0.892           & 22.15      &0.231       & 0.129       \\

Bhang et. al \cite{Bahng2018ColoringWW} & 0.912          & 22.99      &0.228       & \textbf{0.127}       \\

Our          & \textbf{0.917} & \textbf{23.27}     & \textbf{0.223}       & 0.133    \\ \hline
\end{tabular}
\label{tab:comparison}
\end{table}

To further validate the effectiveness of the proposed model, we evaluate the quality of the generated images using quantitative metrics as well. We use average PSNR, SSIM \cite{wang2004image}, LPIPS(vgg) \cite{zhang2018unreasonable,simonyan2015very} and LPIPS(sqz) \cite{iandola2016squeezenet} measures to compare the similarity of the generated images with the ground truth. As shown in Table \ref{tab:comparison}, the proposed algorithm outperforms the SOTA algorithms in SSIM, PSNR, LPIPS(vgg) measures.

\section{Conclusions}

\label{sec:concl}
In this paper, we proposed a novel image colorization algorithm that utilizes textual encoding as auxiliary conditioning in the color generation process. We found that the proposed framework exhibits higher color fidelity compared to the state-of-the-art algorithms. We have also demonstrated that the proposed framework can also be used for recolorization purposes by modulating the textual conditioning. Though our framework has produced superior results, we have considered only textual conditioning of foreground objects in this work. In the given setting, though the proposed algorithm outperforms the SOTA methods, as the textual descriptions mostly depict the foreground objects ignoring the backgrounds; our method exhibits less fidelity for the background colors. However, this discrepancy of background color is difficult to detect without the ground truth images, which are not available in most of the real-world applications. Hence, this problem can be resolved by adding additional color descriptions for the background. We also observed that as the textual descriptions define the colors of the objects coarsely, to fill the gaps, the proposed method generates certain colors which are not there in the respective ground truths. Thus, in certain cases, our method produced less colorful backgrounds which establishes the necessity of  a more exhaustive textual description for the grayscale images in the future.

\bibliographystyle{IEEEtran}
\bibliography{IEEEabrv,references}

\begin{thebibliography}{10}
\providecommand{\url}[1]{#1}
\csname url@samestyle\endcsname
\providecommand{\newblock}{\relax}
\providecommand{\bibinfo}[2]{#2}
\providecommand{\BIBentrySTDinterwordspacing}{\spaceskip=0pt\relax}
\providecommand{\BIBentryALTinterwordstretchfactor}{4}
\providecommand{\BIBentryALTinterwordspacing}{\spaceskip=\fontdimen2\font plus
\BIBentryALTinterwordstretchfactor\fontdimen3\font minus
  \fontdimen4\font\relax}
\providecommand{\BIBforeignlanguage}[2]{{%
\expandafter\ifx\csname l@#1\endcsname\relax
\typeout{** WARNING: IEEEtran.bst: No hyphenation pattern has been}%
\typeout{** loaded for the language `#1'. Using the pattern for}%
\typeout{** the default language instead.}%
\else
\language=\csname l@#1\endcsname
\fi
#2}}
\providecommand{\BIBdecl}{\relax}
\BIBdecl

\bibitem{goodfellow2014generative}
I.~Goodfellow, J.~Pouget-Abadie, M.~Mirza, B.~Xu, D.~Warde-Farley, S.~Ozair,
  A.~Courville, and Y.~Bengio, ``Generative adversarial nets,'' in \emph{The
  Conference on Neural Information Processing Systems (NIPS)}, 2014.

\bibitem{Caesar2018COCOStuffTA}
H.~Caesar, J.~R.~R. Uijlings, and V.~Ferrari, ``Coco-stuff: Thing and stuff
  classes in context,'' \emph{2018 IEEE/CVF Conference on Computer Vision and
  Pattern Recognition}, pp. 1209--1218, 2018.

\bibitem{WelinderEtal2010}
P.~Welinder, S.~Branson, T.~Mita, C.~Wah, F.~Schroff, S.~Belongie, and
  P.~Perona, ``{Caltech-UCSD Birds 200},'' California Institute of Technology,
  Tech. Rep. CNS-TR-2010-001, 2010.

\bibitem{Deng2009ImageNetAL}
J.~Deng, W.~Dong, R.~Socher, L.-J. Li, K.~Li, and L.~Fei-Fei, ``Imagenet: A
  large-scale hierarchical image database,'' \emph{2009 IEEE Conference on
  Computer Vision and Pattern Recognition}, pp. 248--255, 2009.

\bibitem{Levin2004ColorizationUO}
A.~Levin, D.~Lischinski, and Y.~Weiss, ``Colorization using optimization,'' in
  \emph{SIGGRAPH 2004}, 2004.

\bibitem{Huang2005AnAE}
Y.-C. Huang, Y.-S. Tung, J.-C. Chen, S.-W. Wang, and J.-L. Wu, ``An adaptive
  edge detection based colorization algorithm and its applications,'' in
  \emph{MULTIMEDIA '05}, 2005.

\bibitem{Bastos2013RUNTIMEGS}
R.~Bastos, W.~C. Wynn, and A.~Lastra, ``Run-time glossy surface self-transfer
  processing,'' 2013.

\bibitem{anwar2020ColorSurvey}
S.~Anwar, M.~Tahir, C.~Li, A.~Mian, F.~S. Khan, and A.~W. Muzaffar, ``Image
  colorization: A survey and dataset,'' \emph{arXiv preprint arXiv:2008.10774},
  2020.

\bibitem{Wang2018GeneratingHQ}
P.~Wang and V.~M. Patel, ``Generating high quality visible images from sar
  images using cnns,'' \emph{2018 IEEE Radar Conference (RadarConf18)}, pp.
  0570--0575, 2018.

\bibitem{Tola2008AFL}
E.~Tola, V.~Lepetit, and P.~V. Fua, ``A fast local descriptor for dense
  matching,'' \emph{2008 IEEE Conference on Computer Vision and Pattern
  Recognition}, pp. 1--8, 2008.

\bibitem{Perazzi2016ABD}
F.~Perazzi, J.~Pont-Tuset, B.~McWilliams, L.~V. Gool, M.~H. Gross, and
  A.~Sorkine-Hornung, ``A benchmark dataset and evaluation methodology for
  video object segmentation,'' \emph{2016 IEEE Conference on Computer Vision
  and Pattern Recognition (CVPR)}, pp. 724--732, 2016.

\bibitem{Zhang2016ColorfulIC}
R.~Zhang, P.~Isola, and A.~A. Efros, ``Colorful image colorization,'' in
  \emph{ECCV}, 2016.

\bibitem{Carlucci2018DE2CODD}
F.~M. Carlucci, P.~Russo, and B.~Caputo, ``$(de)^2co$: Deep depth
  colorization,'' \emph{IEEE Robotics and Automation Letters}, 2018.

\bibitem{Cheng2015DeepC}
Z.~Cheng, Q.~Yang, and B.~Sheng, ``Deep colorization,'' \emph{2015 IEEE
  International Conference on Computer Vision (ICCV)}, pp. 415--423, 2015.

\bibitem{Bahng2018ColoringWW}
H.~Bahng, S.~Yoo, W.~Cho, D.~K. Park, Z.~Wu, X.~Ma, and J.~Choo, ``Coloring
  with words: Guiding image colorization through text-based palette
  generation,'' in \emph{ECCV}, 2018.

\bibitem{Kingma2015AdamAM}
D.~P. Kingma and J.~Ba, ``Adam: A method for stochastic optimization,''
  \emph{CoRR}, vol. abs/1412.6980, 2015.

\bibitem{Maas2013RectifierNI}
A.~L. Maas, ``Rectifier nonlinearities improve neural network acoustic
  models,'' 2013.

\bibitem{Mikolov2013EfficientEO}
T.~Mikolov, K.~Chen, G.~S. Corrado, and J.~Dean, ``Efficient estimation of word
  representations in vector space,'' in \emph{ICLR}, 2013.

\bibitem{Ioffe2015BatchNA}
S.~Ioffe and C.~Szegedy, ``Batch normalization: Accelerating deep network
  training by reducing internal covariate shift,'' \emph{ArXiv}, vol.
  abs/1502.03167, 2015.

\bibitem{nair2010rectified}
V.~Nair and G.~E. Hinton, ``Rectified linear units improve restricted boltzmann
  machines,'' in \emph{ICML 2010}, 2010, pp. 807--814.

\bibitem{wang2018esrgan}
X.~Wang, K.~Yu, S.~Wu, J.~Gu, Y.~Liu, C.~Dong, Y.~Qiao, and C.~C. Loy,
  ``Esrgan: Enhanced super-resolution generative adversarial networks,'' in
  \emph{The European Conference on Computer Vision Workshops (ECCVW)},
  September 2018.

\bibitem{Isola2017ImagetoImageTW}
P.~Isola, J.-Y. Zhu, T.~Zhou, and A.~A. Efros, ``Image-to-image translation
  with conditional adversarial networks,'' \emph{2017 IEEE Conference on
  Computer Vision and Pattern Recognition (CVPR)}, pp. 5967--5976, 2017.

\bibitem{Simonyan2015VeryDC}
K.~Simonyan and A.~Zisserman, ``Very deep convolutional networks for
  large-scale image recognition,'' \emph{CoRR}, vol. abs/1409.1556, 2015.

\bibitem{kingma2014adam}
D.~P. Kingma and J.~Ba, ``Adam: A method for stochastic optimization,''
  \emph{arXiv preprint arXiv:1412.6980}, 2014.

\bibitem{Lin2014MicrosoftCC}
T.-Y. Lin, M.~Maire, S.~J. Belongie, J.~Hays, P.~Perona, D.~Ramanan,
  P.~Doll{\'a}r, and C.~L. Zitnick, ``Microsoft coco: Common objects in
  context,'' in \emph{ECCV}, 2014.

\bibitem{Zhang2017RealtimeUI}
R.~Zhang, J.-Y. Zhu, P.~Isola, X.~Geng, A.~S. Lin, T.~Yu, and A.~A. Efros,
  ``Real-time user-guided image colorization with learned deep priors,''
  \emph{ACM Transactions on Graphics (TOG)}, vol.~36, pp. 1 -- 11, 2017.

\bibitem{wang2004image}
Z.~Wang, A.~C. Bovik, H.~R. Sheikh, and E.~P. Simoncelli, ``Image quality
  assessment: From error visibility to structural similarity,'' \emph{IEEE
  Transactions on Image Processing (TIP)}, 2004.

\bibitem{zhang2018unreasonable}
R.~Zhang, P.~Isola, A.~A. Efros, E.~Shechtman, and O.~Wang, ``The unreasonable
  effectiveness of deep features as a perceptual metric,'' in \emph{The IEEE
  Conference on Computer Vision and Pattern Recognition (CVPR)}, 2018.

\bibitem{simonyan2015very}
K.~Simonyan and A.~Zisserman, ``Very deep convolutional networks for
  large-scale image recognition,'' in \emph{The International Conference on
  Learning Representations (ICLR)}, 2015.

\bibitem{iandola2016squeezenet}
F.~N. Iandola, S.~Han, M.~W. Moskewicz, K.~Ashraf, W.~J. Dally, and K.~Keutzer,
  ``{S}queeze{N}et: {A}lex{N}et-level accuracy with 50x fewer parameters and
  <0.5mb model size,'' \emph{arXiv preprint arXiv:1602.07360}, 2016.

\end{thebibliography}

\end{document}